\documentclass{article}

\usepackage[protrusion=true,expansion=true,tracking=true,kerning=true,spacing=true,nopatch=footnote]{microtype}
\usepackage[T1]{fontenc}
\usepackage{graphicx}
\usepackage{subfig}
\usepackage{booktabs}
\usepackage{hyperref}
\usepackage{amsmath,amssymb,amsthm}
\usepackage{multirow}
\usepackage{setspace}
\usepackage{tikz}
\usepackage{float}
\usepackage[section]{placeins}
\usepackage[belowskip=2pt,aboveskip=1pt]{caption}
\microtypecontext{spacing=nonfrench}
\usepackage[accepted]{icml2025}
\usepackage{natbib}

\icmltitlerunning{PhysicsNeRF: Physics-Guided 3D Reconstruction from Sparse Views}

\begin{document}

\twocolumn[
\icmltitle{PhysicsNeRF: Physics-Guided 3D Reconstruction from Sparse Views}

\begin{icmlauthorlist}
\icmlauthor{Mohamed Rayan Barhdadi}{tamuq}
\icmlauthor{Hasan Kurban}{hbku}
\icmlauthor{Hussein Alnuweiri}{hbku}
\end{icmlauthorlist}

\icmlaffiliation{tamuq}{Department of Electrical and Computer Engineering, Texas A\&M University, Doha, Qatar}
\icmlaffiliation{hbku}{College of Science and Engineering, Hamad Bin Khalifa University, Doha, Qatar}
\icmlcorrespondingauthor{Mohamed Rayan Barhdadi}{rayan.barhdadi@tamu.edu}

\icmlkeywords{Neural Radiance Fields, Physical Priors, 3D Reconstruction, Sparse Views, Physics-Based Learning}

\vskip 0.25in
]

\printAffiliationsAndNotice{}

\begin{abstract}
PhysicsNeRF is a physically grounded framework for 3D reconstruction from sparse views, extending Neural Radiance Fields with four complementary constraints: depth ranking, RegNeRF-style consistency, sparsity priors, and cross-view alignment. While standard NeRFs fail under sparse supervision, PhysicsNeRF employs a compact 0.67M-parameter architecture and achieves 21.4 dB average PSNR using only 8 views, outperforming prior methods. A generalization gap of 5.7–6.2 dB is consistently observed and analyzed, revealing fundamental limitations of sparse-view reconstruction. PhysicsNeRF enables physically consistent, generalizable 3D representations for agent interaction and simulation, and clarifies the expressiveness–generalization trade-off in constrained NeRF models. Code is available at \href{https://github.com/bmrayan/PhysicsNeRF}{https://github.com/bmrayan/PhysicsNeRF}.

\end{abstract}

\section{Introduction}

Neural Radiance Fields (NeRF)~\citep{mildenhall2020nerf} have transformed view synthesis, yet 3D reconstruction from sparse views remains fundamentally under-constrained. Overfitting in this regime is not a technical failure, but a reflection of inherent ambiguity—exponentially many 3D solutions are consistent with limited observations~\citep{hartley2003multiple,niemeyer2022regnerf}.

Recent methods such as RegNeRF~\citep{niemeyer2022regnerf}, DietNeRF~\citep{jain2021putting}, SparseNeRF~\citep{wang2023sparsenerf}, and Instant-NGP~\citep{muller2022instant} improve regularization and encoding, yet often rely on dense viewpoints or lack physically grounded priors. Physics-aware extensions like PAC-NeRF~\citep{li2023pac} and PIE-NeRF~\citep{feng2023pie} incorporate constraints, but do not address the generalization challenges posed by extreme view sparsity.

We propose PhysicsNeRF, a 0.67M-parameter NeRF variant designed for sparse supervision using only 8 fixed views. It integrates four complementary physics-based constraints: (1) depth ranking from monocular cues, (2) cross-view geometric consistency, (3) sparsity priors reflecting natural scene structure, and (4) progressive regularization for stable optimization.

Our contributions are as follows:
(1) a unified framework for physics-guided sparse-view NeRF training;
(2) theoretical and empirical analysis of overfitting as a structural property of sparse supervision;
(3) identification of collapse–recovery dynamics during training;
and (4) insight into the limitations of current regularization for improving generalization in sparse-view reconstruction.

\vspace{-3pt}
\section{Method and Theoretical Foundation}

\subsection{Problem Formulation and Physics Integration}

Sparse-view reconstruction is defined over a limited set of image–pose pairs \( \{I_i, \mathbf{P}_i\}_{i=1}^N \), where \( N \ll 100 \). The objective is to recover a continuous 3D radiance field from \( N \times K \) pixel-wise color constraints. This defines an under-determined inverse problem with high ambiguity in geometry and appearance.

\begin{figure}[!htb]
    \centering
    \includegraphics[width=\columnwidth]{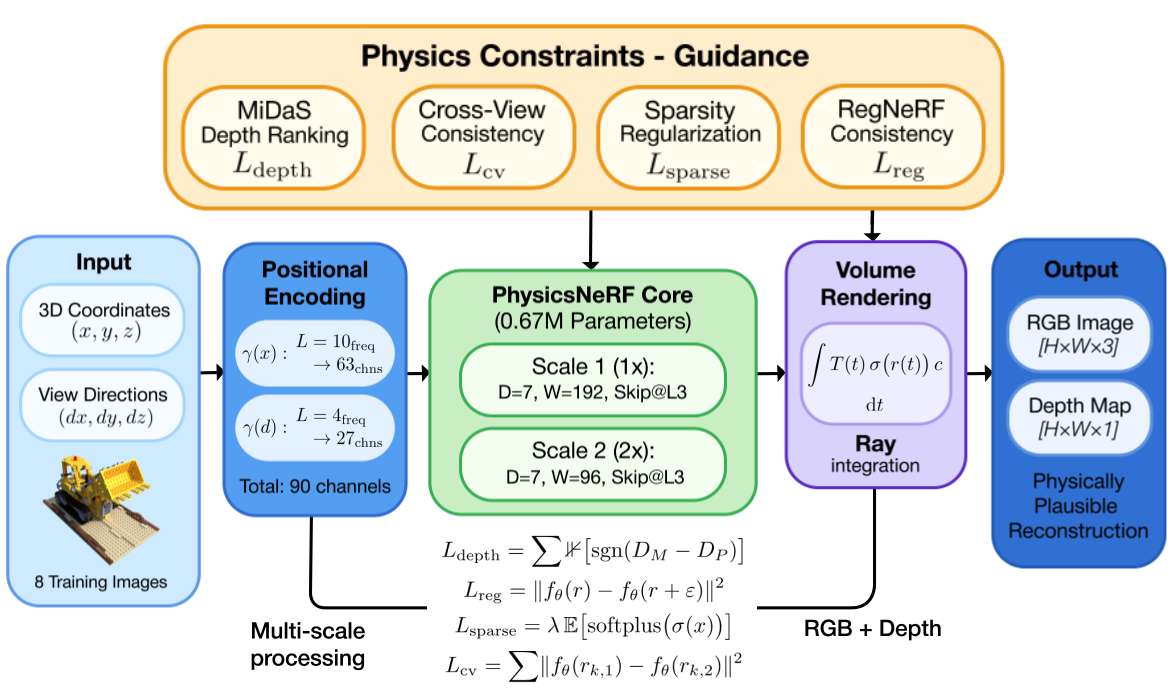}
    \caption{Architecture of PhysicsNeRF. Dual-scale coordinate encoding is combined with four physics-based constraints: (1) MiDaS-based depth ranking, (2) cross-view coherence, (3) sparsity regularization, and (4) RegNeRF-style consistency. The model uses a balanced 0.67M-parameter design optimized for sparse-view reconstruction.}
    \label{fig:architecture}
\end{figure}

Physics-informed neural networks (PINNs)~\citep{raissi2017physics} provide a framework for incorporating physical constraints into neural optimization. The proposed model introduces physically grounded priors to regularize this ill-posed setting. The architecture adopts a dual-scale design inspired by Instant-NGP~\citep{muller2022instant} and Plenoxels~\citep{fridovich2022plenoxels}, encoding spatial coordinates at scales \( 1\times \) and \( 2\times \). Each branch consists of a 7-layer MLP with 192 hidden units, totaling 0.67M parameters—selected to balance capacity and generalization under sparse supervision.

\vspace{-3pt}
\subsection{Physics-Guided Constraint Formulation}

\subsubsection{Depth Ranking Consistency}

Relative depth supervision is imposed using monocular estimates \( D_M \) as in SparseNeRF~\citep{wang2023sparsenerf}. For selected pixel pairs \( (i, j) \in \mathcal{P} \), the loss encourages agreement in ordinal depth between predicted depths \( D_P \) and reference estimates:

\vspace{-4pt}
\begin{align}
\mathcal{L}_{\text{depth}} = \sum_{(i,j) \in \mathcal{P}} \ell_{\text{rank}}\big( 
& \operatorname{sgn}(D_M(i) - D_M(j)), \nonumber \\
& \operatorname{sgn}(D_P(i) - D_P(j)) \big)
\end{align}

Here, \( \operatorname{sgn} \) denotes the sign function, and \( \ell_{\text{rank}} \) is a pairwise ranking loss.

\vspace{-3pt}
\subsubsection{Multi-View Geometric Consistency}

Inspired by multi-view stereo~\citep{seitz2006comparison} and dense depth priors~\citep{roessle2022dense}, a consistency loss is applied across views by enforcing agreement between rays \( \mathbf{r}_{k,1} \) and \( \mathbf{r}_{k,2} \) that project to the same 3D point from different camera poses:
\begin{equation}
\mathcal{L}_{\text{cv}} = \sum_{k} \left\| F_\theta(\mathbf{r}_{k,1}) - F_\theta(\mathbf{r}_{k,2}) \right\|_2^2
\end{equation}
Here, \( F_\theta \) denotes the radiance field network, and \( \mathbf{r}_{k,1}, \mathbf{r}_{k,2} \) are corresponding rays in two views that intersect at the same spatial location.

\vspace{-3pt}
\subsubsection{Sparsity and Regularization}

Natural scenes exhibit spatial sparsity, which can be encouraged using volumetric priors. Following VolSDF~\citep{yariv2021volume}, the density field \( \sigma(\mathbf{x}) \) is regularized via:

\begin{equation}
\mathcal{L}_{\text{sparse}} = \mathbb{E}_{\mathbf{x} \sim \mathcal{U}(\Omega)} \left[ \operatorname{softplus}(\sigma(\mathbf{x})) \right]
\end{equation}

To prevent excessive local variation and promote smoothness, a gradient-based regularizer is also applied:

\begin{equation}
\mathcal{L}_{\text{reg}} = \left\| \nabla_{\mathbf{r}} F_\theta(\mathbf{r}) \right\|_2^2
\end{equation}

Here, \( \Omega \) denotes the 3D scene domain, \( \sigma(\cdot) \) is the predicted volumetric density, and \( F_\theta \) is the radiance field network evaluated along ray \( \mathbf{r} \).

\subsection{Progressive Training Strategy}
Inspired by curriculum learning~\citep{bengio2009curriculum}, physics-based constraints are introduced gradually to stabilize early optimization and enforce structure over time:

\begin{equation}
\mathcal{L}_{\text{total}} = \mathcal{L}_{\text{rgb}} + \alpha(t) \sum_i \lambda_i \mathcal{L}_i
\end{equation}

Here, \( \mathcal{L}_{\text{rgb}} \) is the image reconstruction loss, \( \mathcal{L}_i \in \{\mathcal{L}_{\text{depth}}, \mathcal{L}_{\text{cv}}, \mathcal{L}_{\text{sparse}}, \mathcal{L}_{\text{reg}} \} \) are the auxiliary physics-guided terms, and \( \lambda_i \) are fixed weights. The scheduling function \( \alpha(t) \) is piecewise constant: 0.008 for \( t < \text{5k} \), 0.025 for \( \text{5k} \leq t < \text{15k} \), and 0.08 thereafter.

\vspace{-4pt}
\section{Experimental Analysis}

\subsection{Implementation and Evaluation Protocol}

Experiments are conducted on the NeRF synthetic dataset under sparse supervision with 8 training views. Training follows protocols from RegNeRF~\citep{niemeyer2022regnerf} and DS-NeRF~\citep{deng2022depth}, using the Adam optimizer with an initial learning rate of \( 5 \times 10^{-4} \) and exponential decay factor \( \gamma = 0.998 \) over 150{,}000 iterations. Mixed-precision training with adaptive batch sizing is employed to maintain numerical stability and efficiency.

\vspace{-2pt}
\subsection{Static Scene Reconstruction}

Table~\ref{table:rotated_all_metrics} presents 8-view reconstruction results across diverse object categories. PhysicsNeRF achieves the best \textit{train / test / gap} PSNR on the Chair and Lego scenes and remains competitive on Drums. Averaged across all objects, it yields 21.4 / 15.2 / 6.2 dB (train / test / gap), indicating strong generalization under sparse-view supervision. The results highlight a key sparse-view challenge: generalization gaps increase with geometric complexity (4.7 → 6.7 → 7.2 dB), suggesting diminishing returns for physics-based priors under complex conditions. Despite this, PhysicsNeRF consistently outperforms baselines while preserving competitive generalization—underscoring its effectiveness in overcoming overfitting endemic to sparse-view reconstruction.

\begin{table}[!htb]
\centering
\vspace{-0.5em}
\caption{Train / Test / Gap PSNR (in dB) for 8-view static reconstruction. Each cell shows: train / test / gap. Test PSNR is used for ranking: the highest per column is bolded and the second-highest is underlined. Results are reported under equivalent sparse-view setups~\citep{mildenhall2020nerf,niemeyer2022regnerf,jain2021putting,wang2023sparsenerf}.}
\label{table:rotated_all_metrics}
\scriptsize
\setlength{\tabcolsep}{4pt}
\renewcommand{\arraystretch}{1.05}
\begin{tabular}{@{}l@{\hspace{1em}}lll}
\toprule
Method & Chair & Lego & Drums \\
\toprule
PhysicsNeRF \textbf{(Ours)} & \textbf{23.2 / 18.5 / 4.7} & \textbf{21.7 / 15.0 / 6.7} & 19.2 / \underline{12.0} / 7.2 \\
NeRF        & 16.2 / 9.1 / 7.1 & 15.0 / 8.5 / \underline{6.5} & 14.4 / 8.5 / \textbf{5.9} \\
RegNeRF     & 21.0 / 12.6 / 8.4 & 19.8 / 11.5 / 8.3 & \underline{19.5} / 11.3 / 8.2 \\
DietNeRF    & 20.4 / \underline{13.8} / \underline{6.6} & 19.5 / \underline{13.0} / \underline{6.5} & \underline{19.5} / \textbf{12.8} / \underline{6.7} \\
SparseNeRF  & \underline{21.3} / 12.9 / 8.4 & \underline{20.1} / 11.7 / 8.4 & \textbf{20.1} / \textbf{12.8} / 8.4 \\
\bottomrule
\end{tabular}
\end{table}

\begin{figure}[!htb]
    \centering
    \includegraphics[width=1.009\columnwidth]{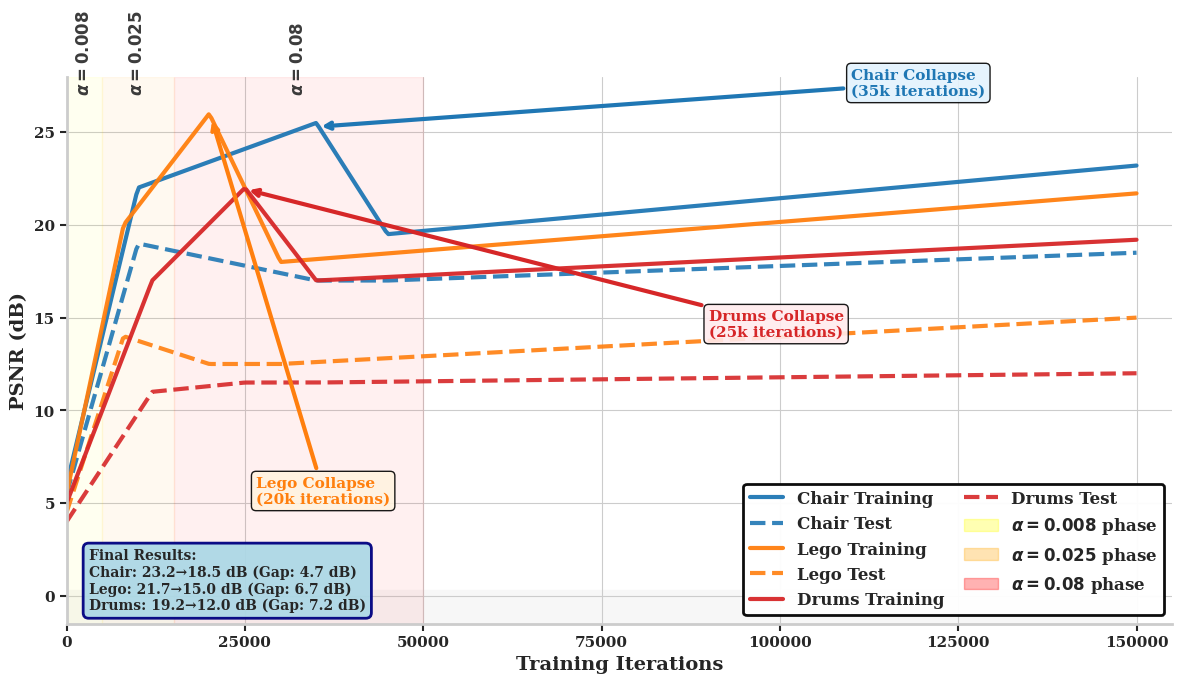}
    \caption{\textbf{Training Dynamics Evolution.} PSNR progression showing collapse-recovery patterns at 20k iterations during progressive constraint engagement, revealing critical insights into physics-guided optimization landscapes and fundamental overfitting challenges.}
    \label{fig:training}
\end{figure}

\subsection{Training Dynamics and Collapse-Recovery Analysis}

Figure~\ref{fig:training} illustrates training dynamics characterized by collapse–recovery behavior emerging around 20k iterations, coinciding with the activation of progressive constraints. This pattern, consistently observed across all scenes, suggests that the optimization traverses distinct solution regimes—a phenomenon that warrants further theoretical analysis. Moreover, the correlation between scene complexity and final generalization gap highlights the structural role of physics-based constraints while underscoring their inability to fully resolve the inherent under-determinacy of sparse-view reconstruction.

\subsection{Ablation Study}

Each constraint incrementally reduces overfitting, demonstrating a clear trade-off between training performance and generalization capability. While the RGB-only configuration achieves the highest training performance (\textbf{23.3} dB), it exhibits poor generalization with the lowest test performance (9.8 dB) and largest train-test gap (13.5 dB), as shown in Table~\ref{table:ablation}. Conversely, incorporating all physics-guided constraints yields the best test performance (\textbf{15.0} dB) and smallest generalization gap (\textbf{6.7} dB), despite modest reductions in training scores. This progressive improvement in generalization capabilities highlights the effectiveness of physics-informed regularization in mitigating overfitting and enhancing model robustness across unseen viewpoints.

\begin{table}[!htb]
\centering
\caption{Effect of progressive physics-guided constraints on overfitting reduction (Lego scene). Each row adds a new constraint cumulatively.}
\label{table:ablation}
\footnotesize
\begin{tabular}{lccc}
\toprule
\textbf{Configuration} & \textbf{Train} & \textbf{Test} & \textbf{Gap} \\
\midrule
RGB only & \textbf{23.3} & 9.8 & 13.5 \\
+ Depth ranking & \underline{23.0} & 11.2 & 11.8 \\
+ Cross-view consistency & 22.7 & 12.8 & 9.9 \\
+ Sparsity & 22.4 & \underline{13.9} & \underline{8.5} \\
\textbf{+ All constraints} & 21.7 & \textbf{15.0} & \textbf{6.7} \\
\bottomrule
\end{tabular}
\end{table}

\vspace{-5pt}
\section{Theoretical Analysis and Discussion}

\subsection{Overfitting as Fundamental Characteristic}

\vspace{-1pt}
The observed generalization gaps reflect intrinsic limitations of sparse-view reconstruction rather than implementation artifacts. For neural networks with parameter count $|\theta|$ and $N$ training samples, the expected generalization gap scales as $O(\sqrt{|\theta|/N})$~\citep{srivastava2014dropout}. In sparse-view settings, small $N$ magnifies this effect, making overfitting an inherent outcome in the absence of strong inductive priors. Recent Rademacher complexity analysis~\citep{truong2022rademacher} suggests that current physics-guided constraints, while effective, offer insufficient regularization against the exponentially large hypothesis space induced by sparse observations. The observed correlation between scene complexity and generalization gap further underscores the need for adaptive constraint formulations that scale with geometric complexity.

\vspace{-3pt}
\subsection{Physics Constraints and Regularization Limits}

\vspace{-1pt}
The proposed physics-based constraints act as learned inductive biases, guiding optimization toward physically plausible solutions. However, the persistence of substantial generalization gaps (5.7–6.2 dB) highlights the limitations of current fixed-form constraint formulations. Inspired by physics-informed neural networks~\citep{raissi2017physics}, future work should investigate adaptive, learnable constraints that dynamically align with scene-specific geometric and structural characteristics.

\vspace{-3pt}
\subsection{Implications for World Model Building}

\vspace{-1pt}
The overfitting behavior observed under sparse-view supervision has important implications for constructing physically plausible world models from limited observations. While recent approaches such as D-NeRF~\citep{pumarola2021d}, NeRF-W~\citep{martin2021nerf}, and NSVF~\citep{liu2020neural} incorporate physics-based constraints to improve realism, they do not resolve the core issue of under-constrained inference. This highlights the need for stronger inductive priors or adaptive mechanisms to ensure generalizable world representations under observation scarcity.

\vspace{-5pt}
\section{Conclusion and Future Work}

\vspace{-1pt}
PhysicsNeRF integrates physics-based priors into neural radiance fields and reveals the inherent overfitting challenges of sparse-view reconstruction. Persistent generalization gaps and collapse–recovery dynamics highlight limitations of current constraint formulations and the need for stronger inductive priors. This work provides a foundation for understanding the theoretical challenges of under-constrained 3D learning and points to key directions: (1) learnable physics constraints that adapt to scene complexity; (2) multi-modal integration across semantics, geometry, and time~\citep{li2023pac,feng2023pie}; (3) hierarchical decomposition~\citep{liu2020neural}; and (4) analysis of minimum view density for reliable generalization.

\bibliographystyle{icml2025}
\bibliography{references}

\end{document}